%% file: main.tex
\documentclass[fleqn,twocolumn, 10pt]{wlscirep}
\usepackage[utf8]{inputenc}
\usepackage[T1]{fontenc}
\usepackage[switch]{lineno} % Line numbers

\usepackage{pifont}
\usepackage{placeins}
\usepackage{amssymb}
\usepackage{amsmath}
\usepackage{url}
\usepackage{xcolor}
\usepackage{verbatim} % Multi-line comments
\usepackage{adjustbox} % Scaled tables
\usepackage{multirow} % Multirow tables

\definecolor{newcolor}{rgb}{.8,.349,.1}
\definecolor{LightGray}{gray}{0.95}
\definecolor{codegreen}{rgb}{0,0.6,0}
\definecolor{codegray}{rgb}{0.5,0.5,0.5}
\definecolor{codepurple}{rgb}{0.58,0,0.82}
\definecolor{backcolour}{rgb}{0.95,0.95,0.92}

\title{What is wrong with Continual Learning in Medical Image Segmentation?\\  \large Moving beyond catastrophic forgetting and towards practical knowledge accumulation.}
\author[1,*]{Camila Gonz\'alez}
\author[1]{Nick Lemke}
\author[1,2]{Georgios Sakas}
\author[1]{Anirban Mukhopadhyay}
\affil[1]{Technical University of Darmstadt, Karolinenpl. 5, 64289 Darmstadt, Germany}
\affil[2]{MedCom GmbH, Dolivostraße 11, 64293 Darmstadt, Germany}
\affil[*]{camila.gonzalez@gris.tu-darmstadt.de}

\begin{abstract}
Continual learning protocols are attracting increasing attention from the medical imaging community. In continual environments, datasets acquired under different conditions arrive sequentially; and each is only available for a limited period of time. Given the inherent privacy risks associated with medical data, this setup reflects the reality of deployment for deep learning diagnostic radiology systems. Many techniques exist to learn continuously for image classification, and several have been adapted to semantic segmentation. Yet most struggle to accumulate knowledge in a meaningful manner. Instead, they focus on preventing the problem of \emph{catastrophic forgetting}, even when this reduces model plasticity and thereon burdens the training process. This puts into question whether the additional overhead of knowledge preservation is worth it -- particularly for medical image segmentation, where computation requirements are already high -- or if maintaining separate models would be a better solution. We propose \emph{UNEG}, a simple and widely applicable multi-model benchmark that maintains separate segmentation and autoencoder networks for each training stage. The autoencoder is built from the same architecture as the segmentation network, which in our case is a full-resolution nnU-Net, to bypass any additional design decisions. During inference, the reconstruction error is used to select the most appropriate segmenter for each test image. Open this concept, we develop a fair evaluation scheme for different continual learning settings that moves beyond the prevention of catastrophic forgetting. Our results across three regions of interest (prostate, hippocampus, and right ventricle) show that UNEG outperforms several continual learning methods, reinforcing the need for strong baselines in continual learning research.
\end{abstract}
\begin{document}

\flushbottom
\maketitle

\section{Introduction}

Supervised deep learning in a static setup is considered the de-facto standard for benchmarking the performance of learning-based medical image segmentation systems. In the static setup, an annotated dataset is divided into training, validation and testing subsets onto which the performance of the learning system is evaluated. Before making this division, all available data is shuffled to ensure that the samples are identically distributed. Yet this is not realistic for diagnostic radiology \cite{pianykh2020continuous}. To obtain robust medical imaging models, it is necessary to leverage data from a variety of sources and continue learning over time. But two constraints may arise when handling medical data due to privacy regulations. These are that (a) data must be stored in predefined servers and so multiple datasets cannot be shuffled, and (b) some of it is only available for training the model during a limited period.

\begin{figure}[t]
\centering
\includegraphics[width=9cm]{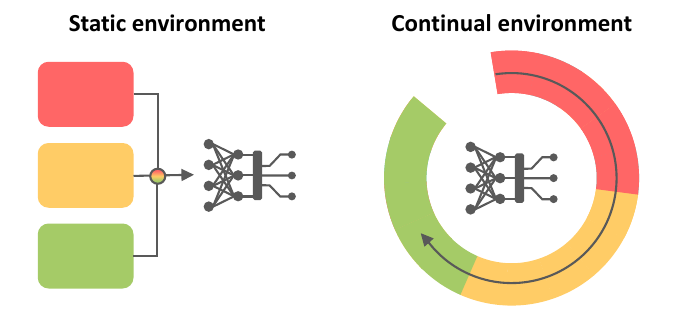}
\caption{Static vs. continual protocols for training and evaluating machine learning models. Each color represents a different domain. In static settings, all data is shuffled and used at once to train the model. In continual environments, the model acquires knowledge over time and can cope with losing availability to part of the data.}\label{figure:learning_paradigms}
\end{figure}

The medical imaging community is becoming aware of the discrepancies between how deep learning algorithms are evaluated and the performance drop that occurs in real clinical settings. This manifests in an increased interest in alternative training and evaluation protocols such as federated \cite{brisimi2018federated,silva2019federated,li2020federated} and continual learning \cite{baweja2018towards,mcclure2018distributed,van2019towards,lenga2020continual,matsumoto2020continual,ozgun2020importance,ozdemir2018learn,ravishankar2019feature,venkataramani2019towards,elskhawycontinual,hofmanninger2020dynamic}. Continual learning in particular addresses the possibility that datasets from different domains arrive sequentially and are only accessible during a certain time interval, as illustrated in Figure \ref{figure:learning_paradigms}.

The main technical challenge of continual learning is the prevention of \emph{catastrophic forgetting}, which occurs when deep learning models adapt too strongly to idiosyncrasies in the last data batches and lose the ability to handle domains seen in the initial stages of training. Most existing continual learning approaches look to prevent this loss, and their evaluations focus on this aspect. Yet the actual goal of continual learning should be achieving \textit{positive backward transfer}, which occurs when the performance on data from early domains \emph{improves} as training continues. In a real clinical setting, only an approach that is successful in this second goal would be deployed. Otherwise, maintaining a separate model for each data source would be preferred.

Particularly for the problem of medical image segmentation, where large model architectures are used and computational requirements are already high due to the dimensionality of the data, we find multi-model approaches to be a practical solution. Of course, we do not always have access to \emph{domain identity labels} during inference. That is to say, we may not have any knowledge on the origin of a particular image. The question is then raised of how to select the most appropriate model. For this, we propose a simple approach based on image autoencoders. By training an autoencoder network per task, we can select the segmenter corresponding to the lowest reconstruction error. 

The \emph{Expert Gate} method has been previously proposed for image classification using autoencoders that reconstruct features extracted from an \emph{AlexNet} network \cite{aljundi2017expert}. Considering the architectural similarities between autoencoder and segmentation networks, in that both produce an output with the same dimensionality as the input, we instead suggest \emph{replicating the segmentation architecture for the autoencoder}, and merely adapting the last layer and training objective to the mean squared error. This poses the advantages that (1) the architecture is already optimized for the problem at hand, (3) the process of data preparation and pre-processing, which has likewise been tuned for the specific problem, can remain the same and (3) no additional design decisions are required. We refer to this multi-model solution as the \emph{U-Net Expert Gate}, or \textbf{UNEG}. We show that this is a better strategy for the segmentation problem with our results across three different regions of interest (ROIs), namely prostate, hippocampus, and right ventricle, using the state-of-the-art \emph{nnU-Net} pipeline \cite{isensee2021nnu}.

Despite several strategies being introduced to permit continual learning, no article has, as of yet, properly introduced the variants and related terminology of continual learning in the medical imaging segmentation context and proposed an \textbf{evaluation scheme that moves beyond forgetting prevention} for each setting. This makes it difficult to compare different approaches and assess their potential usability in clinical practice. A major goal of this article is to provide a holistic view regarding the trade-offs and terminologies of existing strategies, as well as the differences between continual learning scenarios. We propose fair multi-model benchmarks to compare against new continual learning approaches that take these into consideration. In this way, we aim to establish a common ground to facilitate discussion around continual learning in the coming years.

Our contributions are as follows:

\begin{enumerate}

\item Introducing a fair multi-model benchmark for continual learning in medical image segmentation; a solution that can be easily used alongside highly specialized models. The method uses an \emph{oracle} to select the appropriate model in situations where (a) domain identify information is not provided or (b) the system should handle observations from previously-unseen sources.
\item Proposing an autoencoder-based oracle that follows the same architecture as the segmentation network, which is already suitable for the current image modality and region of interest, thereby requiring no additional design decisions.
\item Showing the effect of catastrophic forgetting in three magnetic resonance imaging (MRI) segmentation tasks, namely prostate, hippocampus, and right ventricle, and how this can be avoided.

\end{enumerate}

We start this work by formalizing the problem of continual learning and describing existing scenarios in section \ref{sec:problem_form}. We give an overview of related work on continual learning and its applicability to image segmentation and medical data in section \ref{sec:related_work}. In section \ref{sec:methods}, we describe our proposed evaluation setup based on a multi-model approach and a domain identification strategy with image autoencoders. We report results for three different MR segmentation problems in section \ref{sec:results}. Finally, we give an outlook for future research in section \ref{sec:outlook}.

\section{Problem formulation} \label{sec:problem_form}

We start this section by introducing key terminology and a taxonomy for continual learning settings based on a) how data distributions from different sources differ and b) whether domain identity information is available during inference. We then motivate the need for specialized continual learning solutions to prevent catastrophic forgetting. Finally, we propose a new way to evaluate continual learning approaches that moves beyond forgetting prevention.

\subsection{Continual Learning}

In a continual learning setting, the database consists of $N$ datasets $\left\{ X_i, Y_i\right\}_{i \le N}$. Each dataset comprises samples $( x, y )$ where $x$ is an input and $y$ is the corresponding annotation. In this work we focus on image segmentation, so each sample is a pair of an image $x$ and segmentation mask $y$. The goal is to train a model $\mathcal{F}$ that performs well on all datasets, which arrive sequentially and are only available for a limited period. Figure \ref{figure:learning_paradigms} illustrates how data is received in a continual learning protocol. Model $\mathcal{F}$ is trained sequentially with $\left\{ X_1, Y_1\right\}$, $\left\{ X_2, Y_2\right\}$ and so on until it has acquired information contained in all available data sources. Unlike in a static setup, samples $( x, y ) \in \left\{ X_1, Y_1\right\}$ share certain characteristics resulting from the generation process which are not present in samples from other training stages.

\begin{figure}[h!]
\centering
\includegraphics[width=9cm]{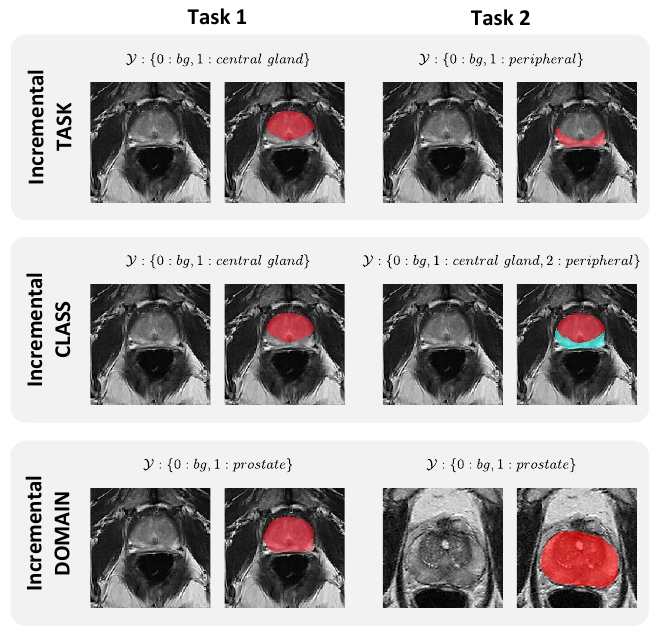}
\caption{Three continual learning scenarios, exemplified for the case of prostate segmentation. In the incremental task scenario, both the image and label spaces can widely differ, but task identity information is available during inference. The incremental class problem consists of sequentially expanding the number of classes that a model can handle. Finally, in the incremental domain setting, the label space remains the same but there are differences in the image space, often resulting from the acquisition protocol, and we have no access to domain identity labels.
} \label{figure:learning_scenarios}
\end{figure}

Differences in the sample distributions can manifest in various ways. van de Ven et al. \cite{van2022three} introduce three continual learning scenarios. These are demonstrated in Figure \ref{figure:learning_scenarios} for the case of prostate segmentation on T2-weighted MRI.

In the \textit{incremental task scenario}, both the input and label spaces can vary, and usually task identity information is provided. For the exemplary task, the network would segment the prostate central gland for task 1, and the peripheral zone for task 2. This scenario is not seen frequently for image segmentation. In the \emph{incremental class scenario}, new classes are incrementally added to the label space. In our example, Whilst only the class for the central gland can be learned with dataset 1, an additional class for the peripheral zone is learned with dataset 2, and the final model can segment both classes. This scenario is interesting to explore, but only meaningful under certain specific circumstances, and adapting the architecture is required each time a new class is introduced. Finally, in the \emph{incremental domain scenario}, the knowledge learned is the same semantically, but there are differences in terms of image characteristics, that is to say that $X_i \simeq X_j$ and $Y_i = Y_j$. We argue that this is the most prevalent scenario in practice for medical imaging, as it comes into play each time a system must learn incrementally from images acquired from different sources and/or at different times. We thus focus on this scenario for the rest of this work.

A \emph{domain} $\mathcal{D}$ is a set of image characteristics that are particular to the acquisition source but independent of the content of interest. For instance, images obtained with one MR machine may have a different contrast than those obtained with another, and the ROI may be captured from a slightly different angle depending on the acquisition protocol.

We further differentiate between whether domain identity information is available during inference, as this is ambiguous in the incremental domain scenario. Based on this, we define three settings. In the simpler case, that we name \textit{Domain Knowledge}, test inputs have the form $( x, i )$, where $i$ specifies that $x \in X_i$. This would be the case if, for instance, a model is trained with data from three different scanners, and we receive at inference time metadata on the scanner used to acquire each image. In the second scenario, \textit{No Domain Knowledge}, no such information is available during testing. The main advantage of having domain identity information is that \emph{domain-specific parameters} can be maintained that are not shared across domains, so only a subset of the model parameters, the \emph{shared parameters}, must be trained in a continual fashion. For classification, the feature extraction part of the model is typically shared whereas the last network layers are domain-specific and set during inference depending on the domain precedence of the test instance.

There are two limitations on the usability of the \textit{Domain Knowledge} scenario. Firstly, the model can only be applied to data from domains that have been observed during training. Secondly, it is not realistic to assume that domain labels will be available during deployment. There are a lot of variabilities in how information regarding image acquisition is encoded in the metadata of image files, even within the same institution. In certain cases, such as in teleradiology systems, this information may not be available due to the anonymization process. Additionally, in a realistic dynamic setting where multiple factors vary over time, it is not trivial to assess which of those factors is the most relevant.

\subsection{The Catastrophic Forgetting debacle}

The na{\"i}ve way to train a model continuously is to perform sequential fine-tuning, executing training steps as data arrives. But if samples belonging to distinct datasets stem from different distributions, this violates the assumption that data be i.i.d. as required by stochastic gradient descent. One prevalent degree of variability for diagnostic radiology are particularities in the image domains that arise from the acquisition protocols or equipment vendors used for each dataset, a phenomenon commonly known as \emph{domain interference} or \textit{domain shift} \cite{glocker2019machine}.

Neural networks trained using stochastic gradient descent on sequentially arriving data adapt too strongly to domain properties present in the last batches. For data similar to that seen in the initial stages of training, this causes a significant drop in performance known as \textit{catastrophic forgetting} \cite{mccloskey1989catastrophic}. If, instead, the model is protected sot hat it does not change too much, it is possible that future knowledge cannot be acquired. Therefore, special attention must be taken during training to ensure that the final model performs well on data similar to that seen at \emph{any stage of training}. However, we argue that simply preventing forgetting and ensuring model plasticity are only the first objectives of continual learning.

\subsection{Existing approaches to mitigate Catastrophic Forgetting and their evaluation}

Different strategies have been developed to reduce the degree of forgetting. The applicability of several popular continual learning methods to medical imaging has also been explored in the past, both for classification \cite{kim2018keep,lenga2020continual,ravishankar2019feature,hofmanninger2020dynamic} and segmentation \cite{ozdemir2018learn,mcclure2018distributed,van2019towards,venkataramani2019towards,ozgun2020importance,matsumoto2020continual}. These methods are compared against the baseline of performing sequential fine-tuning and other continual learning approaches, as well as against the upper bound of performing static training that would be preferred if all data were available at once.

However, we find that often these methods are not compared against the simple multi-model benchmark, i.e. maintaining one model per domain, which could potentially outperform the proposed approach. They \textit{are} compared to the static setup of training a model with all, \emph{joint}, data. Yet unlike a multi-model solution, this is a clear upper bound that is not applicable in a continual learning setting.

In the \textit{Domain Knowledge} case, where domain information can be used at test time, a na{\"i}ve way to avoid catastrophic forgetting is to maintain a separate model $\mathcal{F}_i$ for each domain $\left\{ X_i\right\}_{i \le N}$. Model $\mathcal{F}_1$ is only trained with data $\left\{ X_1, Y_1\right\}$, $\mathcal{F}_2$ is initialized with the parameters of $\mathcal{F}_1$ and fine-tuned with $\left\{ X_2, Y_2\right\}$, and so on. Of course, only model $\mathcal{F}_N$ leverages all available data, as it is sequentially fine-tuned with all data batches. However, no catastrophic forgetting takes place. Each model $\mathcal{F}_i$ maintains the same performance it had after training on data $\left\{ X_i, Y_i\right\}$. Additionally, models do not suffer from a loss of plasticity as the training process does not discourage large parameter changes.

We found only three methods \cite{karani2018lifelong,ozdemir2018learn,venkataramani2019towards} in medical imaging to report a similar comparison. In all cases, either the multi-model solution is treated as an upper bound or the improvement against it is minimal. In addition, some publications report a \textit{backward transfer} or \textit{forgetting} measure that compares the performance at the end of training with all domains to that at the end of training with the corresponding one \cite{nguyen2020dissecting,ravishankar2019feature,ozgun2020importance,elskhawycontinual}. The results only rarely show a situation of \emph{positive backward transfer} or \emph{negative forgetting}, where the performance improves after learning from other domains.

This is worrying, as it questions the applicability of the proposed methods in real clinical workflows. If a model does not improve after continuing training, what advantage comes in maintaining a single model, other than reduced storage requirements? Particularly in clinical practice, space constraints are rarely an issue, and robustness is always the priority.

Despite this, it is common practice to assume domain identity information during testing.
In the \emph{No Domain Knowledge} scenario or if the model should be applicable to previously-unseen domains, \textbf{many continual learning methods cannot be used directly}.

\subsection{Proposed evaluation / Our solution}

\definecolor{domain_info}{RGB}{102, 173, 162}
\definecolor{domain_info_used}{RGB}{113, 186, 209}
\definecolor{no_domain_info_used}{RGB}{90, 116, 174}

We argue that all methods proposed for continual learning should be compared against a \emph{multi-model benchmark}, as it is a straightforward solution to the catastrophic forgetting problem and thus the trivial lower bound for performance. While other continual learning methods require a particular way of training the model or architectural adjustments, the benchmark does not alter the training procedure or architecture of the main model.

\begin{figure}[t]
\centering
\includegraphics[width=9cm]{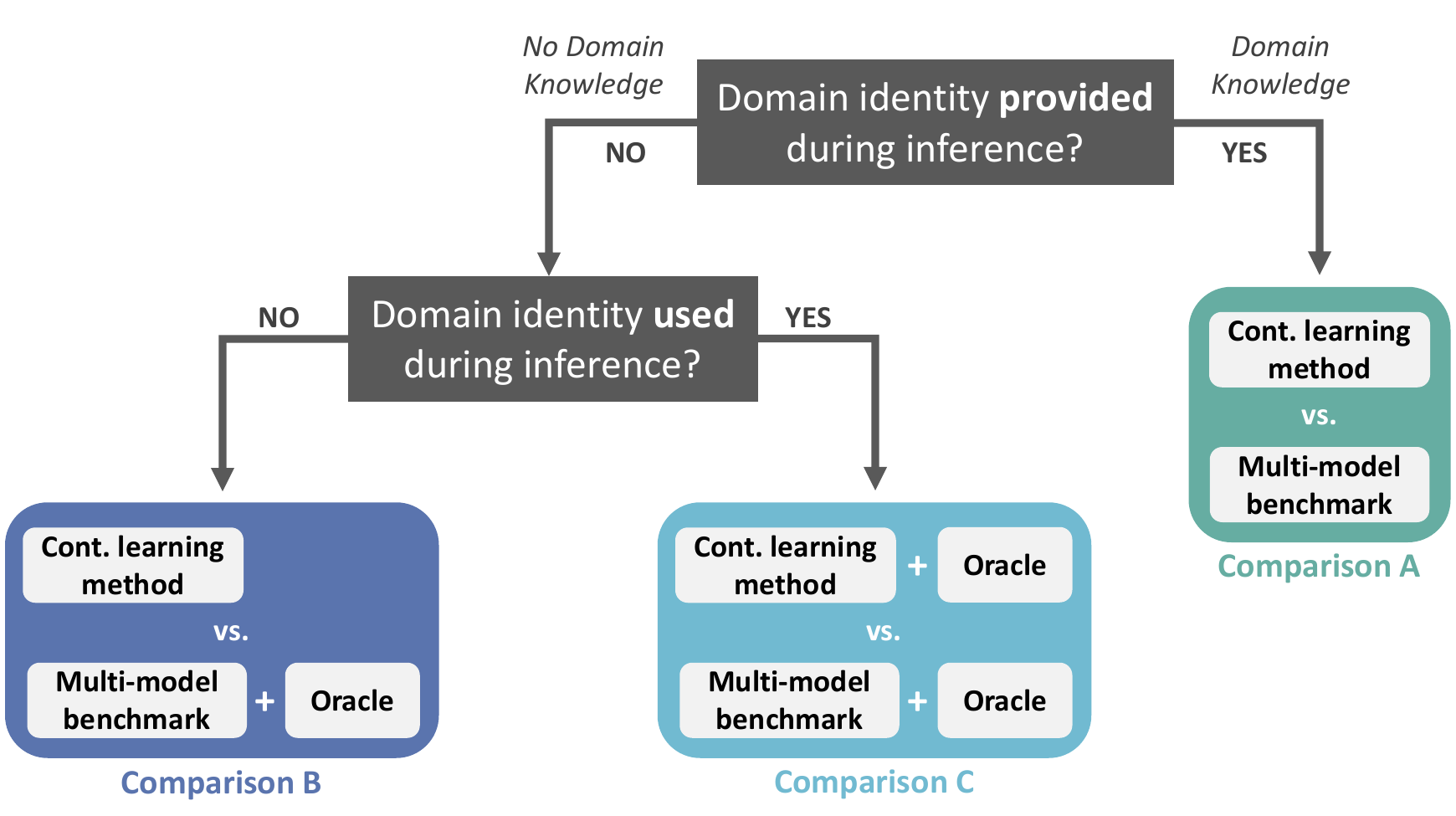}
\caption{Flowchart for the proposed evaluation of continual learning methods. If domain identity information is available, the method should be compared against a benchmark containing one model per domain (comparison A). If domain information is not available, or the model should be usable for data of previously-unseen domains, then it should instead be compared against a combination of the multi-model benchmark and an oracle that identifies which model to apply. If the method requires such information, then the same oracle can be used in conjunction (comparison C). Otherwise, the method is evaluated without using domain knowledge (comparison B). If the benchmark outperforms the newly proposed method, the latter is only an improvement under additional problem-specific constraints, such as limited memory.}\label{figure:intro_flowchart}
\end{figure}

In Figure \ref{figure:intro_flowchart}, we illustrate how an ideal evaluation would proceed. \textbf{\textcolor{domain_info}{If domain identity information can be used during inference}}, a comparison can take place with the benchmark. If domain information is unavailable, we propose using an \emph{oracle} to infer the closest domain for an incoming image $x$. \textbf{\textcolor{domain_info_used}{If the proposed method requires domain information}}, then a comparison can take place for both the continual learning method and the benchmark using domain information inferred by the oracle. \textbf{\textcolor{no_domain_info_used}{If the method does not use domain information}}, then only the benchmark would make use of the oracle.

If the proposed method outperforms the benchmark, the evaluation can go forward. If it performs worse, then its use is only reasonable under additional constraints, such as limitations on persistent memory.

\section{Related work} \label{sec:related_work}

In this section, we give an overview of common continual learning strategies and exemplary methods. We also summarize recent research on continual learning for image segmentation and diagnostic radiology.% (see Table \ref{table:related_work}).

\subsection{Strategies to prevent catastrophic forgetting}

Methods to prevent catastrophic forgetting can be broadly classified into the following strategies.

\textbf{Rehearsal} methods store a subset of examples from previous tasks and periodically interleave these during training \cite{ratcliff1990connectionist,rebuffi2017icarl,aljundi2019gradient}. 
They perform best in practice but do not scale well to an increasing number of domains and can only be used if there are no restrictions against storing training data. \textbf{Pseudo-rehearsal} approaches simulate the same effect without requiring the storage of data files. Within this category are methods that use long-term and short-term memory components \cite{ans2000neural,kemker2017fearnet}, generate examples similar to those of previous tasks with generative models \cite{draelos2017neurogenesis,shin2017continual,ostapenko2019learning,rao2019continual} or use distillation losses to encourage the outputs of the latest model to remain close to previous outputs \cite{li2017learning,lee2019overcoming}. Despite no data being explicitly stored, care should be taken when using generative methods that the models not be sufficient to regain the data \cite{Yu_2019_ICCV}.

\textbf{Sparse-connectivity} methods discourage overlap between representations learned while training with different tasks \cite{2014knoblauch, 2014goodrich, 2015ellefsen}, based on the theory that representational overlap causes catastrophic forgetting. Some methods directly reserve a certain portion of the network for each task. Yet unused network regions are masked, and inference takes place with all regions up to that of the current task \cite{golkar2019continual, mallya2018packnet}. Therefore, the performance does not decrease. A disadvantage of this strategy is that the capacity of the network is limited for features that are not shared, and task identity is always required.

\textbf{Network growing} strategies prevent the loss of model capacity by continuously adding new trainable parameters. Some maintain a single network, to which additional layers or neurons are added as new tasks appear \cite{2015terekhov, wang2017growing}. This is especially successful when complemented by rehearsal training \cite{yoon2018lifelong,draelos2017neurogenesis}. Others train a separate model per task and combine these by merging the parameter values \cite{2017jafari, misra2016cross} or learning connections between the models \cite{rusu2016progressive}. Alternatively, one model state is chosen during inference \cite{aljundi2017expert}. The main disadvantages of network-growing approaches are that the space requirements grow linearly with the number of tasks, and network architectures must be continuously adapted.

\textbf{Regularization} approaches calculate an importance value for each parameter after training a model with data for domain $\mathcal{D}_i$, and penalize the divergence from those parameters, weighted by the importance, when training with data of domains $\mathcal{D}_j:j>i$. Methods differ mainly on how they assess the importance \cite{aljundi2018memory, kirkpatrick2017overcoming, 2017zenke}.

\textbf{Bayesian} methods have also been developed to reduce catastrophic forgetting in Bayesian Neural Networks \cite{ebrahimi2019uncertainty, nguyen2017variational, swaroop2019improving}. The disadvantage is that training networks in a Bayesian manner comprises a considerable time overhead.

Other strategies include learning \textbf{domain-invariant features} \cite{ebrahimi2020adversarial} or maintaining \textbf{different batch normalization parameters} \cite{rebuffi2017learning}.

\subsection{Applications to semantic segmentation}

Most work on continual learning has focused on the classification problem. However, recent research has looked into adapting these strategies for semantic segmentation.

Following the pseudo-rehearsal strategy, a simple approach consists of saving image statistics of previous domains for pseudo-example generation \cite{wu2019ace}. Using a distillation loss, Shmelkov et al. \cite{shmelkov2017incremental} prevent catastrophic forgetting for object detection. Michieli et al. \cite{Michieli2019} expand on this by distinguishing between output and feature-level distillation terms, through results show that considering the divergence of intermediate features rarely improves the performance. Recently, Cermelli et al. \cite{cermelli2020modeling} introduce a distillation loss that takes into account how the proportion of background pixels change across domains and show that this improves segmentation performance.

Beyond pseudo-rehearsal methods, Nguyen et al. \cite{nguyen2020dissecting} propose a regularization-based approach that uses saliency maps as a measure for parameter importance, and Matsumoto and Yanai \cite{matsumoto2020continual} introduce a sparse-connectivity method that learns task-specific masks.

Unlike in classification or regression, where even wrong outputs may seem plausible, semantic segmentation poses the additional challenge that output masks must maintain certain characteristics to resemble the ground truth, such as having a certain number of connected components or adhering to geometric properties. Sequential learning causes the integrity of masks to deteriorate, even if the correct ROI is identified. This is reflected in a greater gap between the performance of static and continual learning results in semantic segmentation.

\subsection{Continual learning in medical imaging}

Several works have explored the applicability of continual learning methods to medical imaging, mostly adapting existing regularization and pseudo-rehearsal strategies to the task at hand. In medical image segmentation, the research is mostly focused on brain MRIs.

Lenga et al. \cite{lenga2020continual} evaluate both \textit{EWC} and \textit{LwF} for the problem of Chest X-Ray lesion classification. \emph{EWC} has also been evaluated for glioma \cite{van2019towards} and white matter lesion \cite{baweja2018towards} segmentation. As is often the case with using this approach, the level of catastrophic forgetting is decreased but learning new domains becomes more difficult. \"{O}zg\"{u}n et al. \cite{ozgun2020importance} also adapt the \textit{MAS} regularization method to brain segmentation. The authors slightly modify how the importance is calculated and normalized, which causes a small improvement over the regular \textit{MAS} implementation.

Ozdemir et al. \cite{ozdemir2018learn} look at how best to select previous examples to prevent catastrophic forgetting using a rehearsal method for the task of segmenting humerus and scapula bones on MR images. A rehearsal method with dynamic memory is also evaluated for the problem of chest CT classification \cite{hofmanninger2020dynamic}. Venkataramani et al. \cite{venkataramani2019towards} explore continual lung segmentation on X-Ray images using a memory component that stores data samples for each target domain.

Karani et al. \cite{karani2018lifelong} mitigate the performance loss in brain MR image segmentation obtained with different scanners or scanning protocols. The method consists of learning a U-Net with shared convolutional layers but domain-specific batch normalization layers; and performs slightly better than training a separate network for each domain. However, it requires data from several domains to be available at once.

Some works focus on learning domain-independent features or learning transformations between feature spaces. Kim et al. \cite{kim2018keep} aim to directly create a domain-independent feature space by maximizing the mutual information between the feature space $Z$ and output space $Y$. This is achieved through minimizing the $L2$ distance between features $z$ and the reconstruction $h(g(z))$, where $G:X \implies Y$ and $h=h^{-1}$. The proposed method outperforms \textit{EWC} and \textit{LwF} in the classification of tuberculosis from chest X-Rays, as well as on \textit{CIFAR10} and \textit{CIFAR100}. Elskhawy et al. \cite{elskhawycontinual} use an adversarial approach to disentangle domain-dependent from domain-independent features. Their method outperforms \textit{LwF} in the incremental class learning setting. Ravishankar et al. \cite{ravishankar2019feature} instead propose using feature transformer networks that turn features extracted for each domain appropriate for using with a following classification network. They show positive results for X-Ray pneumothorax and ultrasound cardiac view classification. However, no comparison takes place between the transformed features and those trained for each task.

Finally, the Bayesian \emph{Distributed Weight Consolidation} method is proposed for performing brain segmentation in a distributed manner for an ensemble of networks trained with Variational Inference \cite{mcclure2018distributed}.

Due to the complexity of the semantic segmentation problem and the geometric differences between ground truth masks from different datasets, methods are mostly evaluated with very similar datasets and only one ROI. In this work, we report results for the segmentation of the prostate, hippocampus, and right ventricle.

\section{Methods} \label{sec:methods}

\begin{figure}[h!]
\centering
\includegraphics[width=9cm]{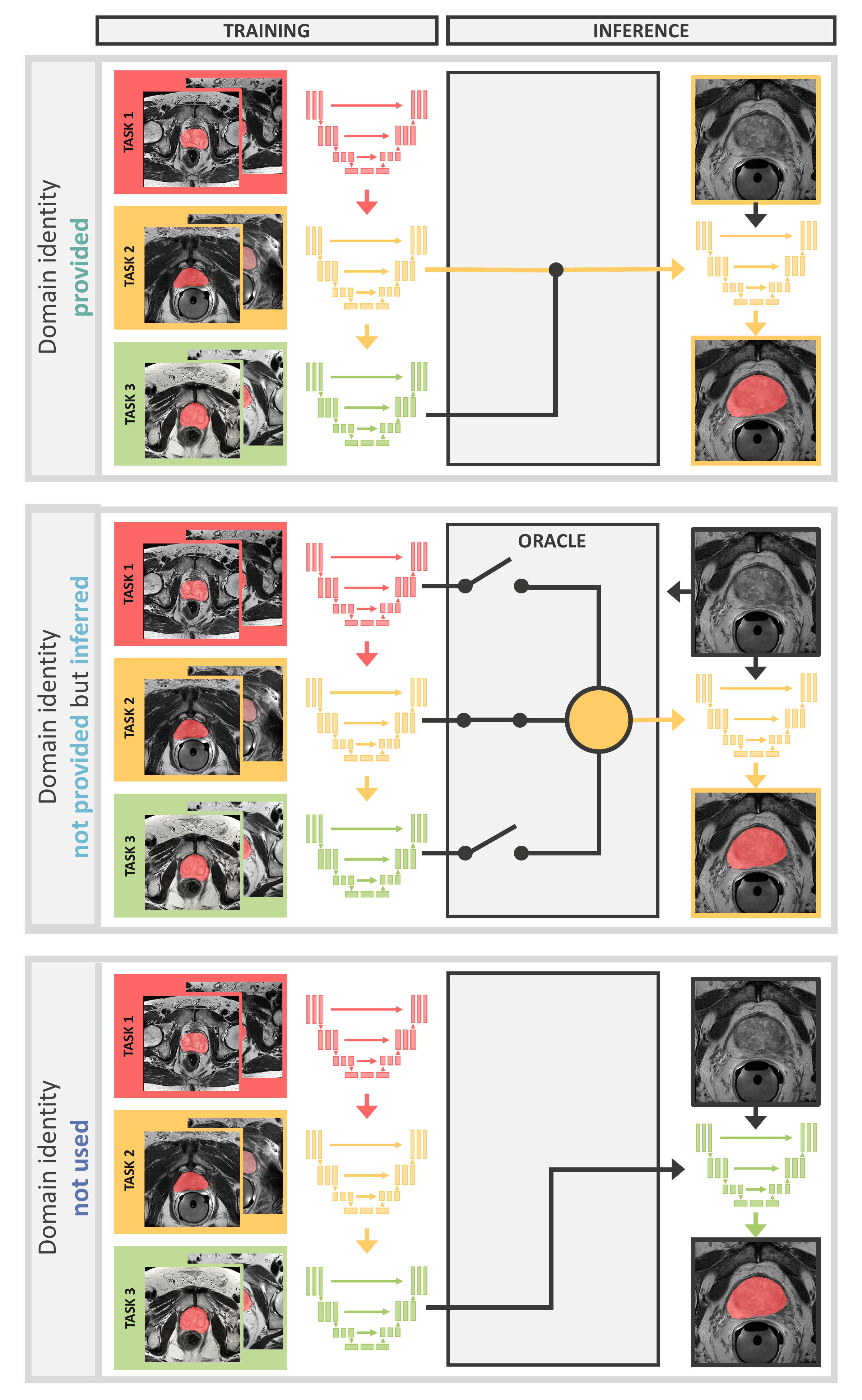}
\caption{Training and testing of continual learning methods under three different settings. During training, shared parameters are trained by sequential fine-tuning, and the state of domain-dependent parameters is saved after each training stage. In the upper image, domain information is provided. During inference, the model state corresponding to the domain of the test image is restored. If this information is not provided, it is inferred through image properties by a domain detection oracle, as shown in the middle diagram. Finally, the lower image shows the case where the continual learning method does not use any domain information.} \label{figure:methods_comp}
\end{figure}

If we store the model state after each training stage, there are three ways to select model parameters depending on whether domain knowledge is provided and/or used. The flowchart in Figure \ref{figure:intro_flowchart} depicts how this translates to different evaluation settings, and Figure \ref{figure:methods_comp} provides a graphical portrayal of how a model state can be selected.

In the \emph{Domain Knowledge} setting where identity labels are given (upper image), the model state right after training with the corresponding domain can be used. If this information is not given, as in the second image, an \emph{oracle} can be used at test time to select the most appropriate model. Finally, single-model methods use the state after finishing training with all domains for extracting all predictions.

Every continual learning method can be compared to a \emph{multi-model benchmark}. As the name suggests, the benchmark maintains one model per domain. It, therefore, poses a fair comparison where no catastrophic forgetting or loss of plasticity take place. In the \emph{Domain Knowledge} scenario, simply using the appropriate model state according to the domain of a test sample is a fair comparison. In \emph{No Domain Knowledge}, the previous method is an upper bound, and an \emph{oracle} must be used to select the best state at test time. 

The main drawbacks of the benchmark are that the space requirement grows linearly with the number of domains and that the model $\mathcal{F}_{i}$ does not leverage information from $\left\{ X_j, Y_j\right\}_{j\geq i}$. There is therefore no possibility of positive backward transfer, though also no forgetting in the individual models.

For the sake of simplicity, we assume that \emph{all} the model parameters are taken from the selected state. In practice, it is possible that only certain layers (such as the end of the decoder) are kept domain-dependent, and the rest are shared. For instance, many continual learning methods maintain separate model \emph{heads}, one per domain.

\subsection{Autoencoder-based oracle}

Inspired by Aljundi et al. \cite{aljundi2017expert}, we use autoencoder networks to build our oracle for domain identification during inference.

An \emph{autoencoder} is a neural architecture designed to reconstruct the input, i.e. learning the mapping $\mathcal{A}:X \rightarrow X$ by minimizing the mean-squared error between the original image and the reconstruction, for $n$ samples (Eq. \ref{eq:mse}).

\begin{equation} \label{eq:mse}
    \mathcal{L}_{MSE} = \frac{1}{n}\sum_{i=1}^{n}(x_i, \mathcal{A}(x_i))^2
\end{equation}

For our oracle, one autoencoder $\mathcal{A}_{i}$ is trained for each incoming dataset. If the hardware allows it, this can occur in parallel to training the segmentation model for the same training stage.

When a test image $x$ arrives, a reconstruction is then created using each of the trained autoencoders $\left \{ \mathcal{A}_{i} \right \}_{i\leq N}$. The domain of the autoencoder with the smallest reconstruction error is used to segment $x$ (Eq. \ref{eq:model_selection}).

\begin{equation} \label{eq:model_selection}
\hat{y} = \mathcal{F}_i(x);\ \mathrm{argmin}_i \; \mathcal{L}_{MSE} (x, \mathcal{A}_i(x))
\end{equation}

Typically, autoencoders contain an \textit{encoder} that reduces the spatial dimensionality of the input and a \textit{decoder} that returns it to the initial dimensions. The method proposed by Aljundi et al. \cite{aljundi2017expert} uses a small CNN to reconstruct \emph{AlexNet} features.

We take a different approach, leveraging the fact that autoencoders follow the same structure as U-Net architectures. We, therefore, propose replicating \emph{the same architecture that is used for the segmentation problem}, as it is already suitable for the specific problem and does not require making any additional design decisions. The only variation we do is that we remove the last layer, which discretizes output values into prediction labels so that we can train the model with the $MSE$ between the inputs and outputs.

\section{Experimental Setup} \label{sec:exp_setup}

In the following, we describe our data corpus for three different image segmentation problems. We also state details on the architecture and training procedure of the segmentation and autoencoder networks and the continual learning methods that we compare.

\subsection{Data}

We evaluate the proposed approach across three anatomies in MRIs, namely the prostate, hippocampus, and right ventricle. 

For prostate, we use T2-weighted MRIs from five different sites \cite{liu2020ms}. The datasets are different in terms of manufacturer and acquisition settings, and each contains 12 to 30 cases. We train in the following order: \emph{BIDMC} $\rightarrow$ \emph{I2CVB} $\rightarrow$ \emph{HK} $\rightarrow$ \emph{UCL} $\rightarrow$ \emph{RUNMC}. The delineations encompass both the central gland and the peripheral zone.

The hippocampus corpus consists of the \emph{Multi-contrast submillimetric 3 Tesla hippocampal subfield segmentation} (henceforth referred to as \emph{Dryad}) dataset \cite{kulaga2015multi}, the \emph{Harmonized Hippocampal Protocol} dataset \cite{wisse2017harmonized} (\emph{HarP} for short) and the data released as part of the \emph{Medical Segmentation Decathlon} (\emph{DecathHip}) \cite{antonelli2022medical}. We train in the order \emph{DecathHip} $\rightarrow$ \emph{Dryad} $\rightarrow$ \emph{HarP}. The segmentation masks cover the posterior and anterior hippocampus.

For right ventricle segmentation, we use the data released for the \emph{Multi-Centre, Multi-Vendor and Multi-Disease (M\&M) Cardiac Segmentation Challenge} \cite{mnm_cardiac_dataset}, which contains two datasets with 75 samples each, the first acquired with \emph{Siemens} scanners, the second with \emph{Philips}.

\subsection{Segmentation nnU-Net}

\label{section:trainingSegmentation}
We use the patch-based, three-resolution variation of the \emph{nnU-Net} \cite{isensee2021nnu}. One model is trained per anatomy, and we perform 250 epochs per dataset.

The architecture and training configuration, such as the patch size, are automatically configured by the framework. As we perform continual training, the settings selected for the first dataset are maintained for subsequent data of the same anatomy. The patch sizes used are $\left [ 28, 256, 256 \right ]$ for the prostate examinations, $\left [ 40, 56, 40 \right ]$ for the hippocampus, and $\left [ 14, 256, 224 \right ]$ for the right ventricle. 

\subsection{Continual Learning baselines}
%\textit{Elastic Weight Consolidation} \cite{kirkpatrick2017overcoming}

%\textit{Learning without Forgetting} \cite{li2017learning}

We compare our proposed benchmark to three popular continual learning methods. We use the implementations and default hyperparameters of the \emph{Lifelong nnU-Net} framework \cite{gonzalez2022lifelong}. All methods are trained in 3D full-resolution with the same configurations as stated in the previous section.

We explore \emph{EWC} (Elastic Weight Consolidation)\cite{kirkpatrick2017overcoming} with a $\lambda_\text{EWC} = 0.4$ , \emph{LwF} (Learning without Forgetting)\cite{li2017learning} with a temperature of $2$ and \emph{MiB} (Modeling the background)\cite{cermelli2020modeling}, with the alpha parameter of $0.9$ and KD loss weighting to $1$. These are the parameters suggested by default. We additionally compare to sequential learning without using any mechanism for knowledge preservation and to the upper bound of static, \emph{joint} training where we use all training data at once.

\begin{figure*}
  \includegraphics[width=\linewidth]{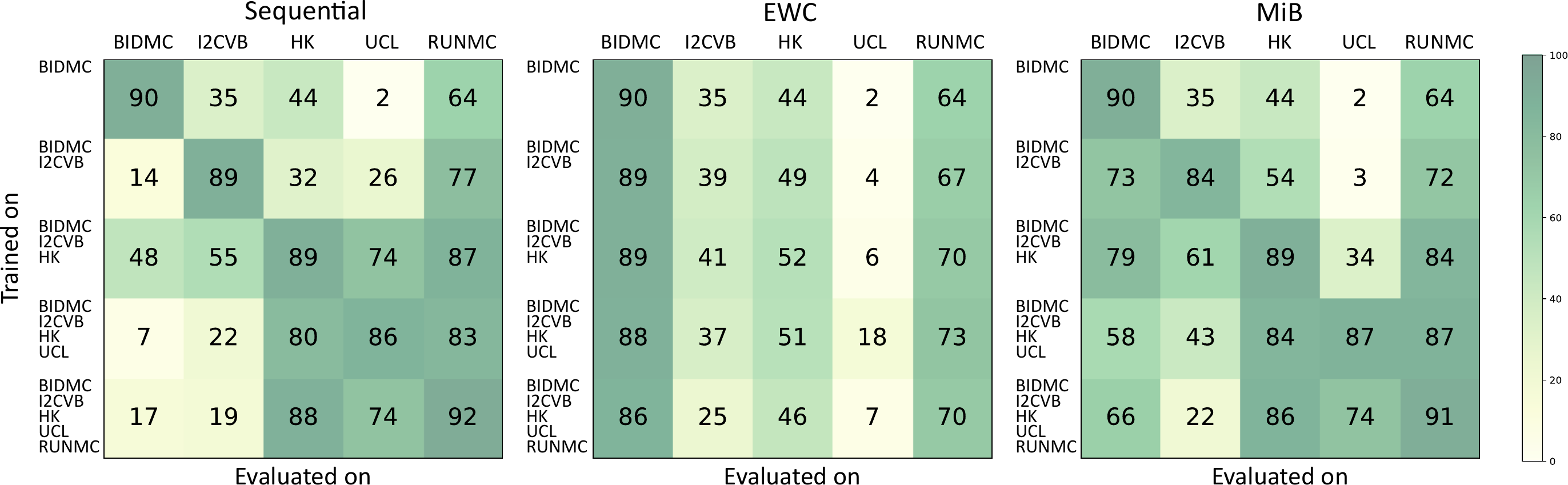}
  \caption{Mean Dice of five prostate segmentation tasks, after training with each of five stages. The diagonal from the upper left to lower right corners shows the score after training with the corresponding training data.} \label{fig:confusion_results_prostate}
\end{figure*}

\subsection{Autoencoder architectures} \label{sec:ae_archs}

The \emph{Expert Gate} method by Aljundi et al. \cite{aljundi2017expert} proposes extracting features from a pre-trained \emph{AlexNet model} \cite{NIPS2012_c399862d} and reconstructing them with a two-layer CNN. We test this approach, though we do not believe it is the most suitable for medical images, and refer to this method as \textbf{AlexNet $\it{z}$-CNN}. Since AlexNet is trained for RBG image classification, we use two-dimensional slices. Replicating the channel 3 times and feeding it through the AlexNet results in a feature volume of 256 channels with reduced spatial resolution.

As autoencoders follow a similar architecture to U-Nets and other popular segmentation models, we propose mimicking the same architecture as used for segmentation. In the case of the nnU-Net, each model is already configured for the specific particularities of the data. We simply modify the last layer to not discretize the logits with a softmax function; and minimize the reconstruction error to the input instead of the segmentation loss. This is our proposed  \textbf{UNEG} \emph{(U-Net Expert Gate)} oracle. We also try an alternative autoencoder that directly reconstructs the input images, namely that offered by the \textbf{MONAI} framework \cite{monaiFramework} which consists of convolution, instance normalization and PReLU blocks. Different to the nnU-Net autoencoder, there is no change in spatial resolution.

To assess whether it is the features or the model which are the most relevant, we experiment as well with using \emph{features from the corresponding nnU-Nets}, reconstructed with a 2-layer CNN autoencoder. We call this method, which also works in three dimensions, \textbf{nnU-Net $\it{z}$-CNN}. Similarly, we try out a \textbf{CNN} network to reconstruct the images directly. In both cases, the features are taken from the last decoder block.

All autoencoders are trained to minimize the MSE between the reconstruction and the input for 250 epochs. During inference/evaluation, we use the segmentation network that corresponds to the autoencoder with the smallest reconstruction error.

\section{Results} \label{sec:results}

We first compare the proposed \emph{UNEG} benchmark to several continual learning methods. We then perform an ablation study where we explore alternatives for the autoencoder oracle. Finally, we look at a few visual examples of reconstructions from the nnU-Net autoencoders.

\subsection{Comparison to continual learning methods}

\begin{figure}[h]
  \includegraphics[width=\columnwidth]{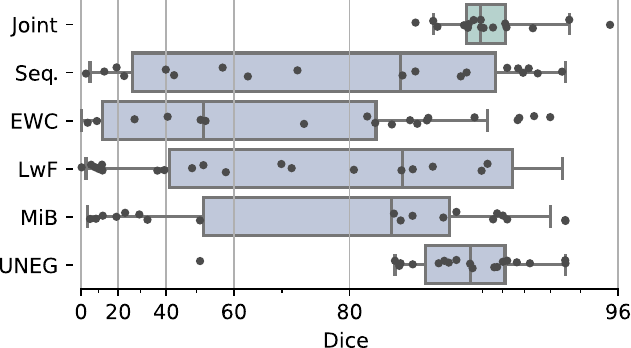}
  \caption{Dice scores of the final model state on test data from five prostate segmentation datasets.} \label{fig:boxplot_results_prostate}
\end{figure}

Figure \ref{fig:boxplot_results_prostate} visualizes the results for prostate segmentation. The first boxplot shows the upper bound of a model trained statically with all data. For prostate segmentation, a Dice of around 90\% is expected in a static scenario. We then see the results for training a model sequentially without forgetting prevention, where the scores are distributed across the performance spectrum. Three continual learning methods follow, namely EWC, LwF and MiB, the latter of which performs best. The multi-model \emph{UNEG} benchmark performs considerably better than continual learning approaches, though there is still a gap in performance to static training.

\input{tables/ablation_study.tex}

Figure \ref{fig:confusion_results_prostate} provides additional insight into how the methods perform at different stages. For regular sequential training, we notice the performance deterioration typical of catastrophic forgetting. \emph{MiB} reduces this effect somewhat, but the final model still produces low-quality segmentations for the first few stages. \emph{EWC} instead displays a different behavior: the performance remains high for the first task, but the model is clearly constrained in its ability to capture new knowledge.

\begin{figure}[h]
  \includegraphics[width=\columnwidth]{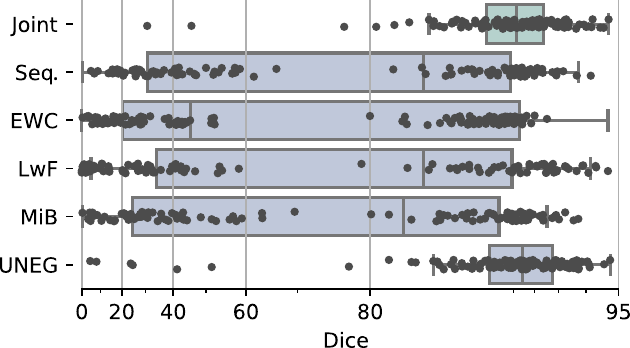}
  \caption{Dice scores of the final model state on test data from the three hippocampus segmentation datasets.} \label{fig:boxplot_results_hippocampus}
\end{figure}

A similar but more pronounced behavior takes place for hippocampus segmentation. Figure \ref{fig:boxplot_results_hippocampus} visualizes these results. For all single-model methods, we see a clear separation between samples that are correctly segmented and those for which performance is dismal. 

\begin{figure}[h]
  \includegraphics[width=\columnwidth]{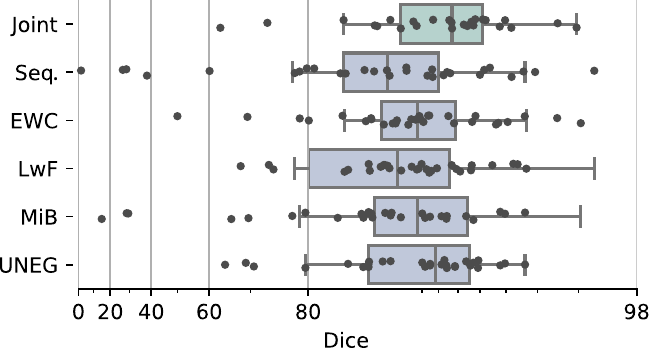}
  \caption{Dice for the task of right ventricle segmentation on test data from both datasets, Philips and Siemens.} \label{fig:boxplot_results_right_ventricle}
\end{figure}

For our third region of interest, the right ventricular blood pool (Figure \ref{fig:boxplot_results_right_ventricle}), all methods perform much better. Only sequentially training the model and \emph{LwF} display a visible performance loss when compared to the joint training upper bound. This is likely due to the fact that there are only two tasks, and the domain differences caused by using different scanners may not be as significant as those introduced in the hippocampus datasets, where the patient populations differ.

\subsection{Ablation study}

We perform an ablation study where we test several autoencoder options in Table \ref{tab:ablation_study}. We calculate the mean Dice over all tasks of the final model states, \emph{BWT} as defined by Gonz\'alez at al. \cite{gonzalez2022lifelong} and how accurately the oracle properly identifies the domain.

The first row shows the upper bound of using the ground truth task identities, as is possible in the \emph{Domain Knowledge} scenario. The second row is the CNN autoencoder proposed by Aljundi et al. \cite{aljundi2017expert}, which reconstructs \emph{AlexNet} features. We then report the results of using the nnU-Net autoencoder (\emph{UNEG}) and several other settings, which are described in Section \ref{sec:ae_archs}. While \emph{AlexNet $\it{z}$-CNN} correctly identifies the domain for most prostate cases, it fails to do so for the hippocampus and cardiac examinations. UNEG instead achieves high accuracy for the three anatomies, but most importantly a high Dice across all domains.

\subsection{Qualitative evaluation of image reconstructions}

\begin{figure*}
\centering
  \includegraphics[width=0.75\linewidth]{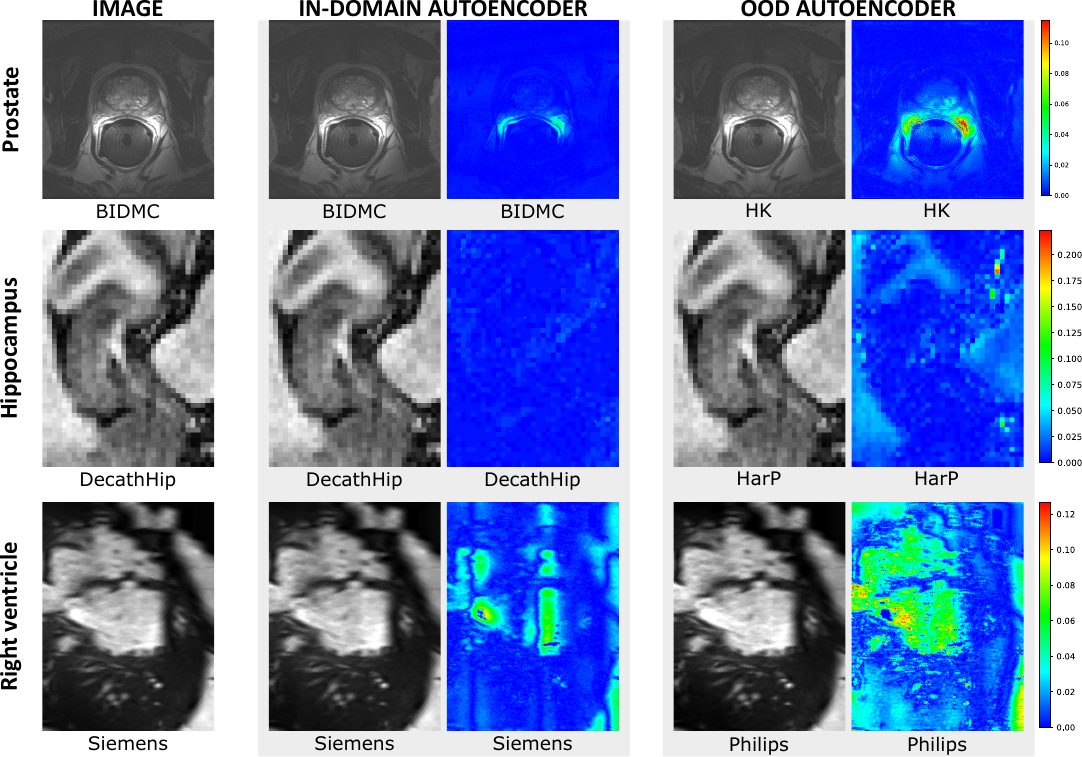}
  \caption{Qualitative evaluation of the reconstructions extracted by the ``correct'' autoencoder vs. an autoencoder trained with a different domain. Besides the reconstructed image we display the residual between the image and the reconstruction.} \label{fig:AE_reconstructions}
\end{figure*}

We can observe exemplary image reconstructions for the three anatomies in Figure \ref{fig:AE_reconstructions}. The first column shows the original image. Then, we see the reconstruction produced by the autoencoder trained with data from the same domain and the residual image highlighting the differences between the two. This is followed by the reconstruction made by an autoencoder from a different domain. The reconstructed images look good at first sight, but when looking at the residual distance, we notice that the quality is much worse than for in-distribution reconstructions.

\subsection{Discussion}

The conversation on continual learning often revolves around \emph{catastrophic forgetting}, the brusque fall in performance for domains seen in early training stages. Often, a trade-off is pursued between \emph{rigidity} and \emph{plasticity}, whereby the model preserves previous knowledge but can still adapt to changes in the environment. Yet there is a trivial way to circumvent the loss of performance in both directions: keeping a separate model per domain -- stored after completing the respective training stage -- and using the appropriate model at test time. 

In this work, we introduce an evaluation workflow alongside a multi-model benchmark, a fair baseline that maintains one model per domain. The benchmark has no possibility of gaining useful information from data seen in later stages for an earlier domain, i.e. no \textit{positive backward transfer} can occur. Yet related works show us that this is rarely the case in practice, which puts into question whether the additional complexity of continual learning approaches that modify the training procedure is worth it.

Central to our evaluation scheme is the question of whether domain identity information is present during inference. Many continual learning methods assume so, which makes it possible to maintain a \emph{multi-head} architecture where only some parameters are shared. Yet this assumption does not hold in many real-world settings, which compromises the applicability of the proposed methodologies.

For cases where identity labels are not known, we propose \emph{UNEG (U-Net Expert Gate)}, which trains one autoencoder per domain. The autoencoder replicates the architecture of the segmentation model -- in our case, a patch-based nnU-Net. This makes use of the fact that the architecture and pre-preprocessing steps are already tuned to the particular input data and task. Our empirical evaluation exploring various autoencoder settings confirms that this is the most effective way to select the correct model at test time.

One limitation of our approach is that one additional model, namely the autoencoder, needs to be trained per stage. This can occur in parallel to training the segmenter but still implies the use of additional computation resources. In future work, we will explore more efficient \emph{oracle} strategies.

\section{Conclusion and Outlook} \label{sec:outlook}

Many methods exist to prevent catastrophic forgetting for image classification, and several have been adapted with relative success to semantic segmentation. Yet \textbf{few methods achieve positive backward transfer}, i.e. while the model does not forget how to deal with data seen in early training stages, it also does not leverage information seen later on. In such cases, a multi-model solution would be preferable in clinical practice, where reliability is paramount and there is rarely a lack of persistent storage. In this work, we present a multi-model strategy alongside a fair evaluation framework for continual learning methods. 

The proposed evaluation considers the fact that continual learning approaches often rely on receiving domain identity information during inference. This may not be the case in real-world dynamic environments, where metadata may be concealed for privacy reasons or the model must handle data from previously-unseen sources.

Continual learning methodologies are gaining a lot of attention from the diagnostic radiology community. Just like we are seeing more works that evaluate models on out-of-distribution data, we hope that training and evaluating models in a continual fashion and quantifying their backward transferability becomes common practice.

\section*{Acknowledgements}

This work was supported by the Bundesministerium für Gesundheit (BMG) with grant [ZMVI1-2520DAT03A].

\bibliography{references}

\end{document}

%% file: tables/ablation_study.tex
\begin{table*}
\caption{Ablation study on the selection of the best autoencoder architecture for an oracle that infers task identity. We report the mean Dice, BWT \cite{gonzalez2022lifelong} and the accuracy at selecting the ``correct'' task identity.}\label{tab:ablation_study}
\centering
\begin{adjustbox}{max width=\linewidth}
\begin{tabular}{l||lll||lll||lll}
&\multicolumn{3}{c||}{\textbf{Prostate}}&\multicolumn{3}{c||}{\textbf{Hippocampus}}&\multicolumn{3}{c}{\textbf{Right ventricle}}\\
Method & Dice $\uparrow$ & BWT (\%) $\uparrow$ & Acc. $\uparrow$ & Dice $\uparrow$ & BWT (\%) $\uparrow$ & Acc. $\uparrow$ & Dice $\uparrow$ & BWT (\%) $\uparrow$ & Acc. $\uparrow$\\
\hline
Task identity	&89.6 \textpm2.0			&-						&-			&90.6 \textpm2.2		&-						&-      &89.6 \textpm1.5		&-						&-     \\
\hline
AlexNet $\it{z}$-CNN		&87.2 \textpm6.6	&-3.5 \textpm6.0		&84.2		&73.3 \textpm19.2		&-23.9 \textpm23.9		&24.1	&81.1 \textpm9.9		&-19.2 \textpm0.0		&50.0  \\
\textbf{UNEG}		&87.0 \textpm6.5			&-3.8 \textpm5.9		&68.4		&89.0 \textpm3.1		&-2.6 \textpm2.6		&97.4	&88.7 \textpm2.3		&-2.0 \textpm0.0		&90.0  \\
MONAI						&66.2 \textpm29.7			&-30.0 \textpm36.2		&31.6		&65.0 \textpm22.8		&-29.4 \textpm29.4		&13.8	&89.6 \textpm1.5		&-0.0 \textpm0.0		&100.0 \\
CNN							&80.3 \textpm11.3			&-11.5 \textpm13.0		&52.6		&72.0 \textpm28.3		&-0.0 \textpm0.0		&56.0	&87.5 \textpm4.3		&-5.6 \textpm0.0		&60.0  \\
%UNet $\it{z}$-MONAI			&73.0 \textpm32.8			&-23.0 \textpm39.8		&84.2		&75.2 \textpm10.5		&-15.0 \textpm14.1		&74.1	&89.6 \textpm1.5		&-0.0 \textpm0.0		&100.0 \\
nnU-Net $\it{z}$-CNN			&84.2 \textpm7.0			&-3.5 \textpm6.1		&52.6		&60.7 \textpm26.4		&-15.0 \textpm14.1		&46.6	&81.1 \textpm9.9		&-19.2 \textpm0.0		&50.0  \\
%AlexNet $\it{z}$-nnUNet		&63.3 \textpm28.0			&-33.2 \textpm33.3		&36.8		&62.4 \textpm24.9		&-14.5 \textpm14.5		&44.8	&81.1 \textpm9.9		&-19.2 \textpm0.0		&50.0  \\
\end{tabular}
\end{adjustbox}
\end{table*}